\documentclass[lettersize,journal]{IEEEtran}

\usepackage[caption=false,font=normalsize,labelfont=sf,textfont=sf]{subfig}
\usepackage[utf8]{inputenc} 
\usepackage[T1]{fontenc}    
\usepackage{hyperref}       
\usepackage{url}            
\usepackage{booktabs}       
\usepackage{amsmath}
\usepackage{amsfonts}       
\usepackage{nicefrac}       
\usepackage{microtype}      
\usepackage{xcolor}         
\usepackage[pdftex]{graphicx} 
\usepackage{wrapfig}
\usepackage{adjustbox}
\usepackage{colortbl}
\usepackage[normalem]{ulem}
\usepackage{soul}

\usepackage{algorithmic}
\usepackage{algorithm}
\usepackage{array}
\usepackage[caption=false,font=normalsize,labelfont=sf,textfont=sf]{subfig}
\usepackage{textcomp}
\usepackage{stfloats}
\usepackage{verbatim}
\usepackage{cite}
\useunder{\uline}{\ul}{}

\newcommand{\TODO}[1]{\textcolor{black}{#1}}

\title{\textbf{NetRoller: Interfacing General and Specialized Models for End-to-End Autonomous Driving}}
\author{Ren Xin, Hongji Liu, Xiaodong Mei, Wenru Liu, Maosheng Ye, Zhili Chen and Jun Ma, \textit{Senior Member, IEEE}
\thanks{Ren Xin, Hongji Liu, Xiaodong Mei, Wenru Liu, Zhili Chen, and Jun Ma are with The Hong Kong University of Science and Technology,  Hong Kong SAR, China (e-mail: rxin@connect.ust.hk; hliucq@connect.ust.hk; xmeiab@connect.ust.hk; wliu354@connect.hkust-gz.edu.cn; zchenei@connect.ust.hk; jun.ma@ust.hk).}
\thanks{Maosheng Ye is with DeepRoute.ai, Shenzhen, China (e-mail: maoshengye@deeproute.ai).}
}
\begin{document}

\maketitle

\begin{abstract}
    Integrating General Models~(GMs) such as Large Language Models~(LLMs), with Specialized Models~(SMs) in autonomous driving tasks presents a promising approach to mitigating challenges in data diversity and model capacity of existing specialized driving models. 
    However, this integration leads to problems of asynchronous systems, which arise from the distinct characteristics inherent in GMs and SMs.
    To tackle this challenge, we propose NetRoller, an adapter that incorporates a set of novel mechanisms to facilitate the seamless integration of GMs and specialized driving models. 
    Specifically, our mechanisms for interfacing the asynchronous GMs and SMs are organized into three key stages.
    NetRoller first harvests semantically rich and computationally efficient representations from the reasoning processes of LLMs using an early stopping mechanism, which preserves critical insights on driving context while maintaining low overhead. 
    It then applies learnable query embeddings, nonsensical embeddings, and positional layer embeddings to facilitate robust and efficient cross-modality translation.
    At last, it employs computationally efficient Query Shift and Feature Shift mechanisms to enhance the performance of SMs through few-epoch fine-tuning.
    Based on the mechanisms formalized in these three stages, NetRoller enables specialized driving models to operate at their native frequencies while maintaining situational awareness of the GM.
    Experiments conducted on the nuScenes dataset demonstrate that integrating GM through NetRoller significantly improves human similarity and safety in planning tasks, and it also achieves noticeable precision improvements in detection and mapping tasks for end-to-end autonomous driving.
    The code and models are available at \url{https://github.com/Rex-sys-hk/NetRoller}.
\end{abstract}


\section{Introduction}

 End-to-End Autonomous Driving~(E2E-AD) models are task-specific Specialized Models~(SMs). They are designed to take raw sensor data as input and generate control commands for operating autonomous vehicles~\cite{uniad,sparsedrive,vad}. 
However, due to the inherent complexity of both perception and decision-making patterns, a substantial amount of labeled sequential rasterized images and control signals is required to ensure robust model performance across diverse scenarios~\cite{tvt_Comp}.
While the labeling-and-training approach can streamline the development and potentially enhance driving similarity to human behavior~\cite{tvt_TrajPlan,tvt_VisionAD}, it incurs high costs for data collection, storage, annotation, and computation. 
Consequently, the generalizability and robustness of a model, which purely learns from sampled demonstrations, would be significantly limited.
Additionally, the model scale is constrained as a trade-off for real-time implementation, which restricts the ability of specialized driving models to capture the relationships between complex driving scenarios and driving behaviors.

\begin{figure}[t]
    \centering
    \includegraphics[width=\linewidth]{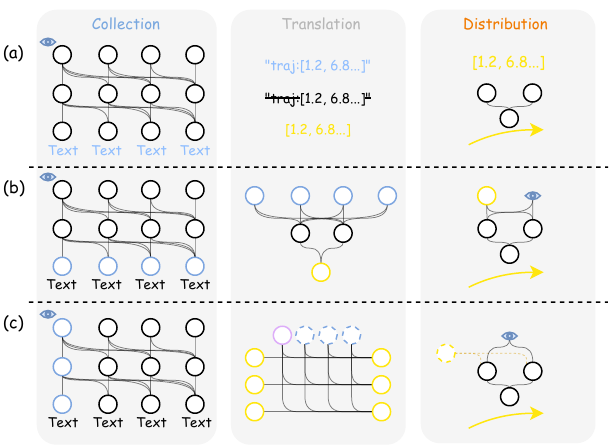}
    \vspace{-2em}
    \caption{
    This diagram illustrates the evolution of the asynchronous frameworks between GMs and SMs. In mode~(a), the traditional approach involves generating textual information using GM, which is then parsed by regular expressions and displayed as received by the fast system. This method relies heavily on predefined patterns and interpretable explicit interfaces.
    Some advanced studies have adopted mode~(b), where selected feature vectors from GMs are inter-modally generated as the latent variables required by SMs. This approach enhances the robustness of information across modalities but still faces challenges in accurately capturing the nuances of the feature vectors and establishing their relevance to SM.
    We propose mode~(c), which fully leverages the latent variables generated during the reasoning process. These latent variables are extracted and robustly translated by a carefully designed module and can be applied to arbitrary targeted feature streams in SM. This method improves the efficiency and accuracy of information transfer, consequently enabling systems to handle complex tasks.
    }
    \label{fig:head}
\end{figure}

Large Language Models~(LLMs) serve as the core foundations of non-task-specific General Models~(GMs).
These foundation GMs are pre-trained on massive corpora of internet and literary text data. 
The inherent properties of textual data, the universal accessibility and exceptional information density, enable highly efficient training at unprecedented scales.
Furthermore, LLMs have less stringent real-time requirements, 
allowing for relatively unlimited model scale and computational complexity.
These advantages enable LLMs to establish a cognitive understanding of commonsense~\cite{nips2020_fewshot}. 
Previous studies have demonstrated that LLMs can outperform humans in certain reasoning tasks, making GMs valuable for applications that need high-level cognition~\cite{simple,gpt4,llama3modelcard,llama2}. 
A comprehensive comparison between GMs and SMs is presented in Table~\ref{tab:summary} to elucidate their distinct characteristics. 

In autonomous driving systems, integrating GMs with specialized driving models offers distinct advantages in multimodal data processing~\cite{tvt_gpt}, yet faces challenges of limited training sample availability and significant computational demands. Notably, recent developments in vision-language models~(VLMs) present promising solutions, demonstrating effective approaches for multimodal learning under such constraints.
VLMs encode visual content into embeddings that are aligned with a smoothed latent space of LLMs, thereby establishing connections between visual and textual modalities~\cite{lavis,blip2,vicuna}.
At its core, this approach implements a general-specialized framework that integrates a dedicated vision captioning model for visual cognition, leveraging GMs' capability of reasoning with textual caption modalities.
The generic reasoning capabilities of GMs enhance the upstream specialized captioning model's capability. 

\begin{table}[t]
    \centering
    \caption{Characteristics of general models and specialized models.}
    \begin{tabular}{c|c|c}
        \toprule
         & General Model~(GM) & Specialized Model~(SM) \\ \hline
        Operation Rate & Lower & Higher \\
        Core Modality & Abstractive language & Specific control value \\
        Scale of Dataset & >100 billion tokens & $\sim$40 kilo frames \\
        Dimensionality & $\sim$4096 & $\sim$256 \\
        Expertise & Human like reasoning & Subconscious action \\
        \bottomrule
    \end{tabular}
    \label{tab:summary}
\end{table}

To facilitate cooperation between GMs and downstream SMs, GMs generate descriptive or instructive replies, which SMs then use as higher-level conditions to guide their actions.
Existing research endeavors in the design of information passing interface have predominantly focused on several distinct paradigms: 
    \textit{(1)~Outputting structured language} to explicitly trigger specialized model~\cite{senna,emma,omnidrive,alphadrive,drivemlm,drivevlm,drivegpt4} requires specialized training data construction and fine-tuning to ensure that the output conforms to a specific format, typically a predefined behavior or trajectory attribution list. However, the formatted output of VLMs is not stable and may include hallucinations. If the model generates non-structured outputs, it can lead to incorrect decisions or even cause program crashes in downstream models.
    \textit{(2)~Utilizing reasoning space tokens as instruction} to downstream models~\cite{vlme2e,vlmad,orion} imposes fewer restrictions on format outputs compared to directly using structured language. 
    However, it still requires waiting for the large model to fully generate its response, which introduces high latency before contributing to the downstream model.
    \textit{(3)~Using special token as the interface} for downstream tasks and applying multimodal supervision has also been explored~\cite{lmdrive, asynchronous, opendrivevla, dsdrive}. However, this approach faces fundamental inter-modality generation challenges~\cite{orion, sora}. Specifically, it requires a single set of model parameters to handle cross-modal generation while maintaining text-aligned latent representations in the special token embeddings, which may compromise performance. Current implementations of this strategy have yet to surpass modern specialized models, and optimal methods for representing and utilizing such specialized tokens remain an open question.
In light of the aforementioned challenges of robustness, high latency, and inter-modality generation, it is imperative to develop a robust, efficient, and modality transformable methodology for integrating general and specialized systems. 
Through comprehensive ablation studies and experimental validation, research from Anthropic~\cite{circuit_tracing} has shown that the latent variables produced between transformer layers in LLMs encode semantically rich, high-level, and instructive token-generating commands for token generation.
These upper-layer latent variables may encompass instructions for generating subsequent words, while the tokens in the final layer might primarily consist of superficial expressions intended for visualization purposes. 
Inspired by these findings, we propose \textbf{NetRoller}, an adaptor that leverages a collective of novel mechanisms to interface asynchronous sets of neural networks analogous to ancient rollers, which enables consistent and smooth relative movement between objects.
NetRoller adopts an information collection mechanism that leverages the first token along with latent variables from all layers to reduce the latency of collecting information from the GM. 
To enhance instruction translation robustness, the collected information, together with non-sense and positional layer embedding, is further extracted by a learnable-query-based transformer and projected to downstream tasks.
Through the proposed Query Shift and Feature Shift mechanisms for instruction distributing modules, which feature a plug-and-play design of the translating module, allow for the independent pre-training of GM and SM without the need for special structures or data designs. Furthermore, a few epochs of fine-tuning can significantly enhance the performance of the SM.
Introduction of NetRoller advantages is illustrated in Fig.~\ref{fig:head}.

Our main contributions are summarized as follows:

\begin{itemize} 
\item We introduce NetRoller, a novel adaptor interfacing the GM-SM asynchronous framework.
It formalizes three key stages to efficiently harness the human-like reasoning ability of GMs, thereby enhancing the robustness of driving models.
\item In the information collecting stage of NetRoller, we propose an instantaneous first-token, all-layer collection mechanism that significantly reduces feature collection latency, which leads to timely enhancement of specialized models. 
\item In the information translation stage of NetRoller, we implement Nonsense Embedding and positional layer embedding in the translation model, enabling \sout{robust and} effective information extraction and compression while offering compatible yet low-dimensional features for downstream models. 
\item In the information distribution stage of NetRoller, we explore instruction distribution mechanisms aimed at identifying computationally efficient methods, namely Query Shift and Feature Shift, to enhance the performance of specialized models with few-epoch fine-tuning.
\item The collection of compact and effective baseline model components, which achieves significant human similarity of planning result and safety improvement on the nuScenes~\cite{nuscenes} dataset, is fully open-sourced for validation and comparison.
\end{itemize}

\section{Related Works}
\paragraph{End-to-End Autonomous Driving~(E2E-AD)}

To minimize the necessity for human intervention, researchers attempt to connect functional models by passing latent variables~\cite{uniad, mp3} instead of separately decoding and then re-encoding, as is typically shown in traditional methods. 
This gives rise to the prototype of E2E-AD, which can reduce the consumption of computational resources and avoid potential information loss during the decoding and re-encoding process. 
Another advantage of passing differentiable latent variables is that it enables task-oriented upstream adjustments, such as planning-oriented perception. 
In this way, the entire autonomous driving system becomes more efficient while enhancing the accuracy and human consistency of decision-making.

Several representative E2E-AD models have emerged in the past few years. 
UniAD~\cite{uniad} is one of the pioneering works in this field, proposing a unified framework to handle various tasks in autonomous driving. 
It lays the foundation for subsequent research by demonstrating the feasibility of end-to-end learning for complex driving scenarios. 
Building upon UniAD, VAD~\cite{vad} introduces a vectorized BEV representation to enhance the performance and reduce computational burden. 
VADv2~\cite{vadv2} further improves upon its predecessor by handling the uncertainty in planning by modeling the planning policy as a probabilistic distribution over actions. 
Mid-to-end training, without integration of a perception module, also contributes to the development of E2E-AD.
PLUTO~\cite{pluto} utilizes contrastive data augmentation to manually introduce linkages between different driving behaviors and specific scenarios.
DiffusionDrive~\cite{diffusiondrive} addresses the computational inefficiency inherent in standard diffusion models by introducing a truncated diffusion policy, enabling real-time multi-mode trajectory generation.
In the aforementioned methods, the improvement in performance results in a clearer and more compact representation of the driving context, reducing overfitting through variational inference and data augmentation.
Among these research directions, enhanced context representation techniques and improved variational inference approaches can substantially boost model performance by maximizing the utility of existing datasets.
The integration of data augmentation techniques enables the model to extrapolate beyond the original data distribution, effectively pushing the performance boundaries constrained by the finite dataset.
However, merely relying on data augmentation is hard to further enhance the generalizability of E2E-AD. More importantly, data augmentation heavily relies on human experience and manual adjustments, which limits its scalability.
Recent advances in Artificial General Intelligence~(AGI) have introduced novel approaches for enhancing specialized driving models, prompting increased research into integrating GMs, particularly VLMs, into autonomous driving systems.
\paragraph{VLMs in Autonomous Driving}
VLMs~\cite{lavis, blip2, vicuna} have shown great potential in various applications by leveraging the complementary strengths of visual and textual information. 
These vision cognitive models~\cite{clip} are pretrained on large-scale datasets that contain images with their annotations, allowing them to establish the relationships between visual and linguistic concepts. 
The vision cognitive model is then integrated with LLMs participating in linguistic comprehension and question answering.
This strategy for enabling vision capability in LLMs is powerful and widely adopted.  
However, there is no definitive consensus on effective and rational deployment of large models in the context of autonomous driving tasks~\cite{llm_review_mdpi}.
DriveGPT4~\cite{drivegpt4} is one of the early works that explores integrating autonomous driving capability to VLMs to enhance the interpretability of the driving model and evaluate the driving skill of the GM.
Following this, several other works have further explored the capability of VLMs in autonomous driving tasks. 
EMMA~\cite{emma} is an impactful research framework that takes multi-view images, historical ego states, and high-level intent commands as inputs. It provides formatted future trajectories, 3D object detection, road graph elements, and scene understanding. In addition to images and ego states, OmniDrive~\cite{omnidrive} also incorporates simulated or actual trajectories, delivering scene descriptions, attention objects, counterfactual reasoning, and planning decisions accordingly. Both frameworks demonstrate the comprehensive reasoning and planning capabilities of VLMs.
In contrast, OpenDriveVLA~\cite{opendrivevla} and VLM-E2E~\cite{vlme2e} focus on specific driving tasks. Particularly, OpenDriveVLA generates trajectories by detokenizing a special planning token based on textual descriptions of objective commands and latent BEV information. While this paradigm significantly improves planning accuracy, it lacks the ability for immediate reaction. On the other hand, VLM-E2E encodes textual annotations of driving context generated by VLMs as instructions for SM, demonstrating a more flexible approach to utilizing knowledge from the GM.
However, the high latency challenges associated with token generation and collection in using VLMs remain unresolved. This characteristic precludes instantaneous high-level instruction, thereby diminishing the benefits of integrating GM into the autonomous driving system that requires real-time responsiveness. Moreover, the compression and distribution of features to specialized models continue to represent a promising area for further exploration.
In light of these insights, we propose NetRoller, which significantly alleviates the information collection problem through an early-stop approach. Additionally, NetRoller enhances robustness and efficiency by stabilizing feature compression and transitions between model components.


\section{Preliminaries}
\textbf{VLMs} have served as a transformative paradigm at the intersection of computer vision and natural language processing, exhibiting unprecedented performance in cross-modal semantic comprehension and reasoning.
The reasoning process of VLMs can be mathematically expressed as follows:
\begin{equation}
    \tau_{i+1} = \text{argmax}\prod_{i=1}^{n} P(\tau_i|\tau_{:i},V;\theta),
\end{equation}
where $\tau_i$ is the $i$-th token, $V$ represents the token-independent visual input, and $\theta$ represents the model parameters of multi-layer transformerss~\cite{bert, attention}. 
VLMs map abstractive natural language to an ultra-high-dimensional latent space and present their reasoning results in an autoregressive manner. 
The autoregressive nature of VLMs implies that each token is generated based on the preceding tokens and the visual input. This sequential dependency results in a computational complexity of $O(n^2)$, where $n$ represents the token sequence length, compounded $t$ times during the reasoning process.
As a result, significant latency is introduced in the output of these large models, which prevents downstream models from effectively utilizing their reasoning capabilities at high frequencies.

\textbf{Instruction fine-tuning} is a technique used to adapt VLMs to specific tasks or domains by providing them with additional training data that includes examples of the desired behavior. This process typically involves fine-tuning the model on a smaller dataset that is relevant to the target task, allowing it to learn the specific patterns and structures of the data. Instruction fine-tuning can significantly improve the performance of VLMs in various applications, such as question answering and dialogue systems. By leveraging the existing knowledge of VLMs and emphasizing their understanding through targeted training, instruction fine-tuning enables more effective and efficient use of these powerful models in real-world scenarios.
The mathematical expression of instruction fine-tuning can be expressed as:
\begin{equation}
    \max_{\Delta\theta} \prod_{i=1}^{n} P(\tau_i|\tau_{:i},V;\theta+\Delta\theta), 
\end{equation}
where $\Delta\theta$ represents the additional parameters or residual values introduced during the finetuning process. This allows the model to adapt its reasoning capabilities to better align with the specific requirements of the task at hand, enhancing its performance and applicability.


\textbf{End-to-end driving models} typically employ a single-shot inference paradigm with a fixed input length.
The model learns to map sensor data of the perceived surroundings and historical ego states to control commands, enabling real-time decision-making.
To improve both generalizability and interpretability, modern end-to-end autonomous driving systems employ multi-task supervision that encompasses not only control commands but also intermediate perception and prediction outputs.
This peripheral supervision framework ensures critical environmental representations are preserved throughout the processing pipeline, and planning modules operate on verified, task-relevant features rather than black-box embeddings.
Mathematically, the end-to-end driving model can be expressed as:
\begin{equation}
    \mathcal{T}, \mathcal{P} = f(\mathcal{S}, H| \theta),
\end{equation}
where $\mathcal{T}$ represents the planned future trajectory, $\mathcal{P}$ is peripheral supervision like Bird-Eye-View~(BEV) and semantic detection results of driving context, $\mathcal{S}$ is the input sensor data, $H$ is the historical information in fixed length, and $\theta$ represents the model parameters. 


\textbf{Integration of asynchronous systems} leverages the GM to guide the inference process of downstream SMs. This approach allows downstream models to leverage the reasoning capabilities of GMs while maintaining the computational efficiency of downstream SM. By decoupling the intensive computational demands of GMs from the real-time constraints, it enables more scalable and responsive processing.
The mathematical expression of the integration can be expressed as:
\begin{equation}
    \mathcal{T}, \mathcal{P} = f(\mathcal{S}, H| \theta_{SM}, f_{\text{VLM}}(\tau_{0:I}, V, \theta_{GM})),
\end{equation}
where $f_{\text{VLM}}$ denotes the inference function of GM, $\tau_{0:I}$ indicates the prompt tokens, $V$ represents the visual input and $\theta_{SM}$, $\theta_{GM}$ represents parameters of SMs and GMs respectively. 
In this integrated framework, the probabilistic reasoning module is invoked at a substantially higher frequency than the VLM. The VLM's reasoning outputs are repeatedly employed as conditional inputs during the downstream inference process. Under ideal conditions, these VLM-derived priors enhance the probability of generating both ground-truth trajectories $\mathcal{T}_{GT}$ and peripheral ground truth $\mathcal{P}_{GT}$.
To achieve this, the VLM should be capable of providing relevant information on time, thus enhancing the granularity of the VLM provided conditions.
Additionally, the specialized model should ensure that it does not overly rely on the reasoning results of the large model during inference, in order to avoid performance degradation due to the absence of valuable instructions from VLM.



\section{Methodology}

\begin{figure*}[t]
\centering
\includegraphics[width=0.9\linewidth]{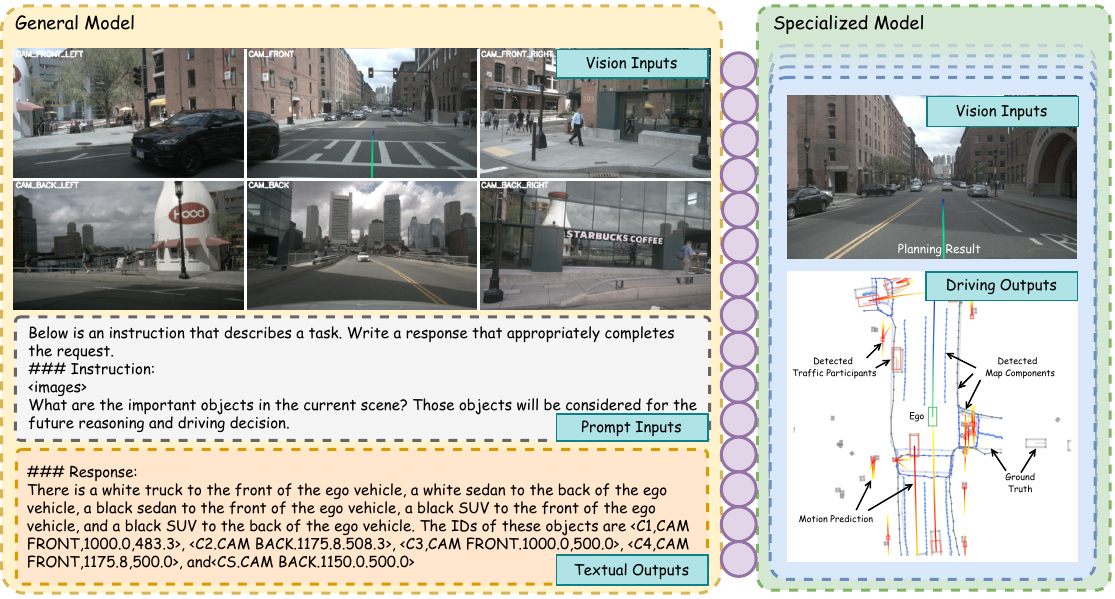}
\vspace{-1em}
\caption{It depicts the input and output modalities of the GM-SM asynchronous system, where the VLM acts as the GM and the E2E-AD model acts as SM. Upon receiving the panoptic vision inputs and prompts, the VLM yields a corresponding response. On obtaining translated instructions from NetRoller~(purple circles), the specialized driving model then undertakes real-time inference grounded in the comprehension of the GM and panoptic visual inputs, thereby generating driving outputs, including detection, BEV, and planning results. The planning results are projected to the vision input for visualization.}
\label{fig:modality}
\end{figure*}


In this section, we elaborate NetRoller, an adaptor equipped with a collective of novel mechanisms for enhancing the performance of an autonomous driving system through asynchronous integration of GM and SM.
A demonstration showing the input-output modality of the asynchronous framework with NetRoller is shown in Fig.~\ref{fig:modality}.
NetRoller introduces innovative mechanisms for collecting and leveraging informative features across both GM and SM. The framework architecture comprises three key components:
\begin{itemize}
    \item \textbf{Information Collection,} which aggregates informative features of the input from GM.
    \item \textbf{Information Translation,} which eliminates the gap of feature modalities from the GM to SM.
    \item \textbf{Information Distribution,} which optimizes feature utilization in SM.
\end{itemize}

In practice, VLMs serve as the GM, which processes raw visual inputs and generates transferable instructions for the specialized driving model to output the detection, mapping, prediction,  and planning results. The complete framework architecture is illustrated in Fig.~\ref{fig:overview}.

\subsection{Information Collection}

\begin{figure*}[t]
    \centering
    \includegraphics[width=1.0\linewidth]{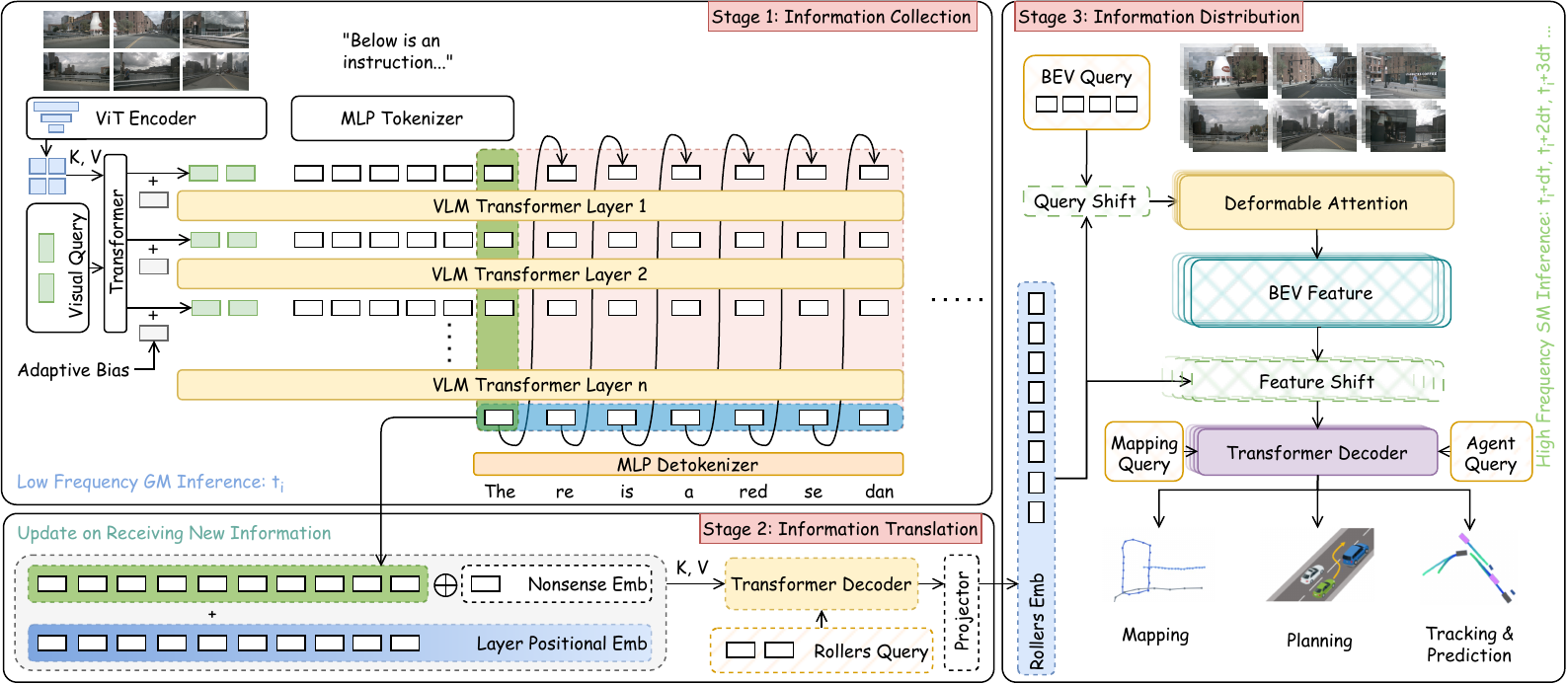}
    \vspace{-2em}
    \caption{\TODO{Overview of GM-SM asynchronous framework with NetRoller. The VLM receives panoptic images and prompts as inputs at time step $t_i$ and generates answer tokens. NetRoller collects the latent instructions and conducts either Query or Feature Shift, instructing the detection and planning modules of the specialized model at the following time steps.}}
    \label{fig:overview}
\end{figure*}

VLMs generate tokens in an autoregressive manner using ultra-high-dimensional feature vectors. This process can be computationally expensive and time-consuming. To address this issue, we propose an information collection mechanism that begins collecting informative features from the very first token. This approach enables more efficient information flow and reduces the computational overhead associated with generating tokens sequentially.
NetRoller collects information from all layers of VLMs, and the VLM-generated latent variables are wrapped with a green dashed box as shown in Fig.~\ref{fig:overview}.
The main principle is to leverage the information generated in layers of transformer computation in a large model while generating text. As indicated by circuit tracing~\cite{circuit_tracing}, the inter-layer information is likely to guide the generation of future tokens. It may contain as much information as the collection of all the last-layer tokens.
NetRoller can optionally continue to collect latent variables of non-initial tokens, thereby minimizing information loss.  
Moreover, although textual responses generated by GM are not mandatory for the driving task in SM, they are preserved to enhance interpretability.


\subsection{Information Translation}
The collected feature vectors contain valuable instructive variables, but they also carry excessively abundant information.
Though necessary for textual outputs, these feature vectors create an extra computational burden when instructing SM.
To reduce the computational burden and project latent variables from GM to the latent space of SM, an information extraction and projection process is conducted. This process reduces the quantity and dimensionality of the collected information while preserving its essential features. 
A lightweight Querying Transformer~(QFormer)~\cite{blip2} equipped with a set of learnable queries, is employed to extract pertinent information from the collected latent variables. 
As the learnable queries in NetRoller serve as crucial parameters for cross-modality translation, we refer to them as "Roller Embeddings", a term that vividly captures their functional role in the asynchronous framework.
The latent variables obtained from the large model are initially identified by \textit{Layer Positional Embeddings}, thereby enabling NetRoller to distinguish the sequence and origin of features effectively. To ensure the stability and robustness of the output and maintain normal operation even in the absence of an upstream model, the processed latent variables are concatenated with a learnable \textit{Nonsense Embedding}, which prevents the occurrence of Not a Number~(NaN) values when utilizing the attention module in scenarios where the upstream model produces no output. This approach also mitigates the risk of propagating hazardous tokens to the specialized system. Ultimately, the extracted variables are projected into the lower-dimensional SM latent space via a feed-forward neural network.
Mathematically, the information compression process can be expressed as
\begin{equation}
    E_R= \texttt{Linear}(\texttt{QFormer}([h_{:i}^{-l:}+PE_{layer}^{-l:}, E_N])),
\end{equation}
where $E_R\in\mathbb{R}^{\text{roller\_num}\times\dim_{SM}}$ represents \textit{Roller Embeddings}, $h_{:i}^{-l:}$ represents last $l$ layer feature vector of the first $i$ tokens,  $PE_{layer}^{-l:} \in \mathbb{R}^{l\times dim_{GM}}$ is the \textit{Layer Positional Embedding}, and $E_N \in \mathbb{R}^{1 \times dim_{GM}}$ is the \textit{Nonsense Embedding}. 
\TODO{To be more specific, \texttt{QFormer} performs cross-attention computations and Multi-Layer Perceptron~(MLP) projections within a single-layer, decoder-only transformer. The resulting feature vectors are then layer-normalized before being passed to the subsequent model.}



\subsection{Information Distribution}

After obtaining the translated information, we need to distribute it to the downstream model for inference. To achieve this, two mechanisms are proposed: Query Shift and Feature Shift, which leverage instructive $E_R$ to shift the stream of latent variables.
\paragraph{Query Shift} The query~($Q\in\mathbb{R}^{n\times dim_{SM}}$) shift mechanism can leverage the decoder only and one layer \texttt{Transformer}~(\texttt{TF}) to shift the query vector to a different position in the latent space. This allows the model to focus on different aspects of the input data, thereby improving its ability to capture relevant information. The Query Shift process can be expressed as:
\begin{equation}
    Q^\star = \texttt{Transformer}( tgt=Q, mem=\text{$E_R$}).
\end{equation}
The Query Shift mechanism enables the query to perform attention calculations on the compressed information one more time, resulting in a new query vector. 
This process extracts important features from the compressed information and provides strong flexibility by ensuring that only relevant queries change, thus maintaining system robustness. 
However, this method introduces a significant computational burden, which may affect the operating frequency of downstream networks.
In addition to shift query vectors with transformer architecture, we also explore applying a uniform bias~(\texttt{Bias}) to query vectors. 
This approach aims to provide a more efficient alternative by introducing a fixed offset to the query vector, which can help guide the attention mechanism without the need for additional attention calculations.
Mathematically, it can be expressed as:
\begin{equation}
    Q^\star = Q + E_R.
\end{equation}
This method transmits the information extraction burden to the upstream QFormer, thereby alleviating the computational pressure on the downstream model. However, this approach may lead to a decrease in the diversity of the shifted query, as all queries are assigned the same bias.
The query bias method requires dedicated learning mechanisms for query adaptation, as it transforms absolute query values into residual representations relative to the upstream model's bias parameters. This residual formulation is essential for maintaining the semantic consistency of the modified queries while preserving their responsiveness to guidance from GM.

\paragraph{Feature Shift} The \texttt{TF} and \texttt{Bias} modes of Feature Shift are similar to those of Query Shift mechanisms, respectively. The mathematical expressions for Feature~($F\in\mathbb{R}^{n\times dim_{SM}}$) Shift mechanisms are as follows:
\begin{equation}
\begin{split}
F^\star &= \texttt{Transformer}( tgt=F, mem=E_R), \\
F^\star &= F + E_R.
\end{split}
\end{equation}

In contrast to Query Shift mechanisms, Feature Shift directly compensates for feature-level deficiencies that the detection module of SM might otherwise fail to capture. While this approach theoretically contributes scenario cognition more directly to SM, it necessitates applying feature adjustments during each downstream inference cycle, inevitably introducing additional computational overhead.

\section{Experiments}

\begin{table*}[ht]
    \centering
    \setlength{\tabcolsep}{4pt}
    \renewcommand{\arraystretch}{1.1}
    \scriptsize
    \caption{Abbreviations of NetRoller configurations.}
    \adjustbox{width=0.65\linewidth}{
    \begin{tabular}{c|c|c}
    \hline
        Parameter & Option & Meaning \\ \hline
        Collection Mechnism & \texttt{C$_0$}, \texttt{C$_1$}, \texttt{C$_{\text{-}1}$} & All token all layer, 1st token all layer, all token last layer \\
        Collection Module & \texttt{PE$_{l}$} & Using layer positional embedding \\
        Freezed Module & \texttt{S$_{f}$, G$_{f}$} & Freezing parameters of SM, GM respectively \\
        Distribution Method & \texttt{TF}, \texttt{Bias} & $E_R$ with transformer, $E_R$ as bias \\
        Distribution Position & \texttt{Q$_{\text{BEV}}$}, \texttt{F$_\text{BEV}$} & Shift BEV query, shift BEV feature \\
        \hline
    \end{tabular}
    }
    \label{tab:abbrev}
\end{table*}

\subsection{Implementation Details}
\paragraph{Dataset} 
The specialized driving model is trained on the nuScenes dataset~\cite{nuscenes}, a widely recognized benchmark for training and evaluating autonomous driving systems. It comprises 1,000 scenarios with diverse driving contexts, including urban, suburban, and highway environments. Each scene is annotated with 3D bounding boxes for various object classes, as well as detailed trajectory information for vehicles and pedestrians. The nuScenes dataset provides a widely adopted comprehensive benchmark for evaluating perception and prediction models in autonomous driving tasks.

We employ the DriveLM~\cite{drivelm} dataset to fine-tune the VLM for adapting the GM to the driving task. The DriveLM comprises 4,871 annotated frames, which are a subset of the nuScenes dataset, each accompanied by descriptions for driving scenario understanding and behavioral instructions. On average, each scenario is annotated with 91.4 QA pairs. Similar to real-world conditions, the driving videos are subsampled, leading to a reduced frame count, while the captions for each frame must encompass a diverse set of factors. This results in a sparse labeling distribution for frames but rich textual annotations, effectively simulating the output behavior of the slow-response GM and unstructured natural language in real-world scenarios\TODO{, which possibly contain misleading or erroneous information}. Additionally, this design mitigates overfitting to a fixed dataset to some extent.



\paragraph{Details of Model}
Our GM builds upon LLaMA-2-7B~\cite{llama2}, a widely-used pretrained large language model that serves as a foundation for comparative studies and downstream development. For visual modality processing, we employ CLIP~\cite{clip}, a popular VLM comprising two key components: an image encoder for visual input processing and a text encoder for linguistic input processing. Both encoders map their respective inputs into a shared embedding space to enable cross-modal understanding.
Through integrated fine-tuning of LLaMA-2 and CLIP on the DriveLM dataset, we get a baseline VLM specifically optimized for autonomous driving applications. This model demonstrates strong performance in driving-oriented visual question answering tasks in perception, prediction, behavior, and planning contexts.
The GM is utilized with a randomly sampled question category together with panoptic image inputs. The reasoning process can be stopped as early as we get the designated iterations of essential latent variables. 

\TODO{In the translating stage, the Layer Positional Embedding is initialized uniformly, while the Nonsense Embedding is initialized randomly. The processed query serves as both the Key and Value in the decoder-only transformer. The multi-head cross-attention module employs 32 heads. Before entering the MLP, layer normalization is applied. The MLP consists of two fully connected layers, with the first doubling the feature dimensionality, followed by a SiLU activation and a normalization layer in between.}

To achieve state-of-the-art end-to-end driving performance, the specialized model requires some key modules:
Deformable Attention~\cite{deformable} is a mechanism that facilitates more flexible and efficient attention computation in neural networks. It enables the model to focus on specific regions of interest in the input data instead of processing the entire input uniformly. This is particularly beneficial in tasks where certain parts of the input are more relevant than others, such as object detection or image segmentation. 
By selectively attending to these regions, deformable attention can reduce computational overhead and enhance performance. 
BEVFormer~\cite{bevformer} is a notable example of a model that employs deformable attention to improve its ability to process and understand complex visual information for BEV driving representation. 
It efficiently extracts various driving-related contexts from multi-view visual inputs. 
Due to the high information density and instructional flexibility of BEV, the Query Shift mode of NetRoller is designed to shift BEV query vectors in each scenario.
In this way, it changes the querying position of visual inputs and boosts the accuracy in detecting valuable objects within a certain scenario.
In contrast to Query Shift, the Feature Shift mode updates the contents in the latent BEV feature vector according to extracted high-level instructions from GM.
We adopt VAD, a framework that incorporates essential perception modules and leverages vector maps as auxiliary supervision to enhance feature recognition and interpretation~\cite{vad}. This approach demonstrates strong performance on driving tasks when evaluated on the nuScenes dataset. Our experiments build upon this baseline to rigorously assess the improvements contributed by our proposed design.


\paragraph{Evaluation Metrics} Evaluations mainly focus on the performance of the perception, mapping, prediction, and planning tasks. 

For perception and mapping, we use the following metrics:
\begin{itemize}
    \item \textbf{NDS}: The nuScenes detection score is evaluated by a weighted sum of mean Average Precision~(mAP) and five True Positive metrics~($\mathbb{TP}$), as
    \begin{equation}
        \text{NDS}=\frac{1}{10} \left[5 \text{mAP}+\sum\nolimits_{\text{mTP}\in \mathbb{TP}}\left(1-\min(1, \text{mTP}) \right) \right].
    \end{equation}
    \item \textbf{Map-mAP}: The mean average precision across all map classes, including divider, pedestrian crossing, and boundary.
\end{itemize}

For prediction and planning, we use the following metrics:
\begin{itemize}
    \item \textbf{Similarity}: The similarity score between the generated and ground truth trajectories, calculated using $L_2$ distance at different time steps~(1\,s, 2\,s, 3\,s).
    \item \textbf{Safety}: The safety score, which measures the percentage of collisions at different time steps~(1\,s, 2\,s, 3\,s).
    Safety should be regarded as the primary indicator because it can simultaneously reflect the improvement in perception and planning capabilities.
\end{itemize}




\paragraph{Training Setup} 
The DriveLM dataset offers QA annotations only for key frames. 
Consequently, during the training phase, the model has to randomly select an annotation from prior frames within the same scene. 
Furthermore, for each key frame, DriveLM's annotation consists of multiple pairs of questions and answers that may have an order relationship, which is also randomly sampled in training.
This random sampling approach aims to simulate potential delays, omissions, and irrelevant outputs that may arise in real-world usage, thereby enhancing the robustness of the GM-SM asynchronous framework.

Model training and inference are conducted on a server equipped with 8$\times$NVIDIA L20 GPUs. The training period lasts for around 8 hours for 5 epochs. The batch size of each GPU is set to 1, but gradients are accumulated for 8 iterations to stabilize the training process. 
We employ a base learning rate of $3.9\times10^{-6}$ per sample~(calculated as 0.001/256) with cosine annealing scheduling. The weight decay is set to 0.02.
Besides, the setup between \texttt{NetRoller-Bias} and \texttt{NetRoller-TF} is slightly different. 
The dimension of $E_R$ for \texttt{Bias} is $1\times256$ while that for \texttt{TF} is $8\times1024$.
The dimension of $E_R$ for \texttt{Bias} is set lower because when applying $E_R$ to specialized models, a $1\times256$ dimensional $E_R$ is efficient for broadcasting and adding to BEV queries, which stands for the extremely simplified distribution strategy. While the BEV feature needs to perform pixel shuffling~\cite{smolvlm} to align with the 1024-dimensional $E_R$ and conduct the flexible, computationally complex but extremely informative value shift with \texttt{Transformer}.
Other abbreviations for model setups are listed in Table~\ref{tab:abbrev}.

\subsection{Overall Performance} 

\begin{figure*}[t]
\centering
    \includegraphics[width=\linewidth]{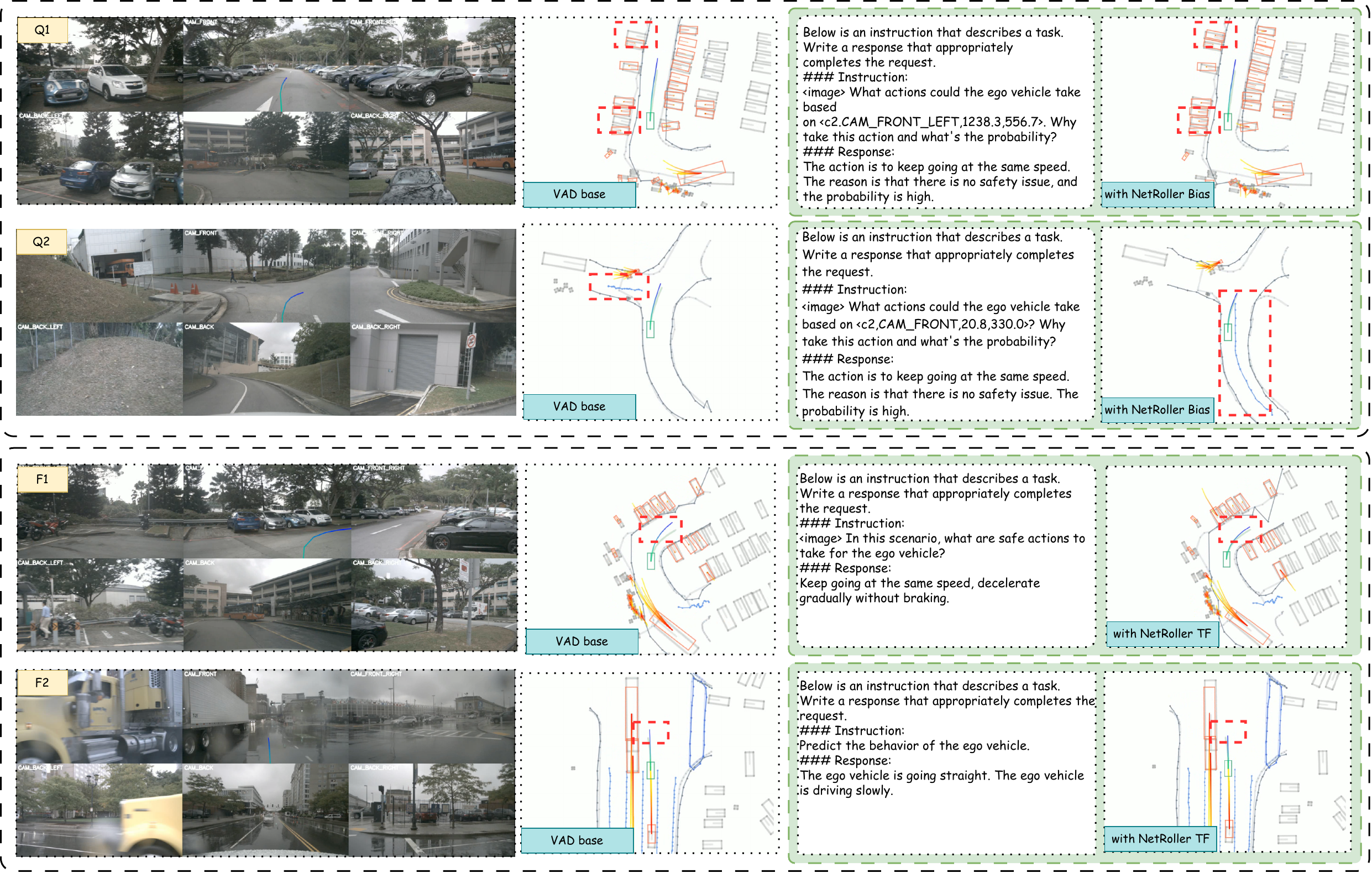}
    \vspace{-2em}
    \caption{\TODO{Qualitative comparisons of perception and planning performance are presented, which evaluate the vanilla VAD Base against its GM-enhanced variants using the NetRoller in \texttt{Bias} mode and \texttt{TF} mode. With GM integrated in NetRoller \texttt{Bias} mode, previously undetected vehicles on the left are successfully captured in scenario Q1. In scenario Q2, the center line—critical for the current planning task—is correctly detected even though it was mistakenly unlabeled in the ground truth. In contrast, when GM is integrated with NetRoller in \texttt{TF} mode, SM produces more human-like trajectories. Specifically, in scenario F1, it is evident that SM plans a more human-like trajectory based on the instructions for longitudinal trajectory planning from the GM; while in scenario F2, the ego vehicle accelerates in a more gradual, human-like manner after yielding to a long vehicle.}}
    \label{fig:qualitative}
\end{figure*}

Qualitative comparisons illustrating the effect of NetRoller are shown in Fig.~\ref{fig:qualitative}. From this, the characteristics of NetRoller in \texttt{Bias} and \texttt{TF} modes, as well as the improvements compared to the vanilla SM, can be directly observed.

\begin{table*}[ht!]
\centering
\setlength{\tabcolsep}{4pt}
\renewcommand{\arraystretch}{1.1}
\scriptsize
\caption{Comparison of visual/language based end-to-end models on the nuScenes dataset.} 
\adjustbox{width=0.75\linewidth}{

\begin{tabular}{c|c|c|cccc}
\hline
                         & Perception      & Map             & \multicolumn{4}{c}{Planning}                                                \\ \hline
                         & NDS$\uparrow$             & mAP$\uparrow$             & $L_2$(m)@1\,s/2\,s/3\,s$\downarrow$ & Avg.$\downarrow$            & Collision(\%)@1\,s/2\,s/3\,s$\downarrow$ & Avg.$\downarrow$            \\ \hline
NMP~\cite{nmp}                      & -               & -               & -/-/2.31       &                 & -/-/1.92               & -               \\
ST-P3~\cite{stp3}                    & -               & -               & 1.33/2.11/2.90 & 2.1100          & 0.23/0.62/1.27         & 0.4250          \\
UniAD~\cite{uniad}                   & -               & -               & 0.48/0.96/1.65 & 1.0300          & 0.05/0.17/0.71         & 0.3100          \\
VLM-E2E~\cite{vlme2e}                  & -               & -               & 1.22/1.94/2.68 & 2.0100          & 0.26/0.60/1.17         & 0.1868          \\
GPT-Driver~\cite{gptdriver}               & -               & -               & 0.27/0.74/1.52 & 0.8400          & 0.07/0.15/1.10         & 0.4400          \\
OpenDriveVLA-7B~\cite{opendrivevla}          & -               & -               & \textbf{0.20}/\textbf{0.58}/1.21 & 0.6600          & 0.05/0.28/0.55         & 0.2900          \\
VAD base~\cite{vad}                & -               & -               & 0.41/0.70/1.05 & 0.7175          & 0.07/0.17/0.41         & 0.2200          \\
\rowcolor[HTML]{EFEFEF} 
VAD base$\star$                & 0.4593          & 0.4786          & 0.41/0.70/1.05 & 0.7175          & 0.07/0.16/0.33         & 0.1900          \\
\hline
NetRoller-Bias-C$_0$    & 0.4612          & 0.4826          & 0.38/0.66/1.01 & 0.6855                & 0.08/0.13/0.30         & 0.1693          \\
NetRoller-Bias-C$_{\text{-}1}$   & 0.4615 & \textbf{0.4838} & 0.39/0.67/1.02 & 0.6964                & 0.10/0.16/0.31         & 0.1901 \\
NetRoller-Bias-C$_1$-PE$_{l}$ & \textbf{0.4700} & 0.4816          & 0.38/0.66/1.01 &         0.6830 & \textbf{0.06/0.13/0.30}       & \textbf{0.1628}          \\ 
\hline
NetRoller-TF-C$_0$      & 0.4572 & 0.4704          & 0.35/0.61/\textbf{0.95} & \textbf{0.6380}       & 0.10/0.17/0.31         & 0.1923 \\
NetRoller-TF-C$_{\text{-}1}$     & 0.4540          & 0.4733 & 0.36/0.63/0.97 & 0.6556                & 0.11/0.17/0.33         & 0.2032          \\
NetRoller-TF-C$_1$-PE$_{l}$   & 0.4544          & 0.4733          & 0.36/0.63/0.98 & 0.6599                & 0.10/0.17/0.33         & 0.1994          \\ 
\hline
\end{tabular}
}
\label{tab:main}

\end{table*}

Performance comparisons against several state-of-the-art models on the nuScenes dataset are presented in Table~\ref{tab:main}. In the table, we compare the performance of several mainstream end-to-end models, including NMP~\cite{nmp}, ST-P3~\cite{stp3}, UniAD~\cite{uniad}, and VAD~\cite{vad}, as well as methods that integrate VLMs, such as VLM-E2E~\cite{vlme2e}, GPT-Driver~\cite{gptdriver}, and OpenDriveVLA-7B~\cite{opendrivevla}.
Table~\ref{tab:main} compares \texttt{Bias} and \texttt{TF} variants of NetRoller under three information collection modes~(\texttt{C$_0$}, \texttt{C$_{\text{-}1}$}, \texttt{C$_1$}), where the GM parameters remain frozen across all configurations~(\texttt{G$_f$}). Note that variants employing the \texttt{C$_1$} collection mode are equipped with \texttt{$PE_{layer}$}, as this component is essential for proper functioning in this mode.

As shown in the table, \texttt{NetRoller-Bias-C$_1$-PE$_{l}$} achieves consistent improvements across all driving metrics compared to our baseline. Specifically, relative to the \texttt{VAD~base$\star$} (tested in-house for perception and mapping performance), it reduces the average collision rate by 16.71\%, demonstrating a significant enhancement in safety. Additionally, it exhibits improved planning similarity.
In contrast, \texttt{NetRoller-TF-C$_0$} excels in planning related tasks, particularly in trajectory similarity, where it achieves a 12.46\% improvement in $L_2$ similarity. However, this configuration shows degraded performance in other operational metrics.
The safety metric demonstrates that in the \texttt{NetRoller-Bias-C$_1$-PE$_{l}$} configuration, the GM enhances real-time feature captioning, thereby improving perception and mapping performance, ultimately leading to safer trajectory generation. Conversely, the similarity metric reveals that in the \texttt{NetRoller-TF-C$_0$} mode, the GM primarily compensates for long-term high-level cognition, producing comparable planning outcomes while neglecting real-time perception and reactive adjustments.

Moreover, variants of NetRoller \texttt{TF-C$_0$} and \texttt{TF-C$_1$-PE$_{l}$} compare information collection modes \texttt{C$_0$} and \texttt{C$_1$} in the \texttt{TF} mode. Our analysis reveals that planning performance using only all-layer feature vectors from the first token achieves comparable results to utilizing full "all-layer all-token" features, suggesting the first token's embeddings effectively capture the essential planning information. By comparing the variants \texttt{NetRoller-TF-C$_1$-PE$_{l}$} with \texttt{NetRoller-TF-C$_{\text{-}1}$}, it is evident that using "all last layer tokens" results in a greater degree of deterioration compared to using "all tokens from the first layer". A similar conclusion can also be drawn by comparing \texttt{Bias} series. This aligns with the hypothesis that intermediate layer feature vectors may contain more useful subconscious thinking high-dimensional information, while the last layer's feature vectors are primarily used for text interaction.
Comparative analysis of NetRoller variants reveals distinct advantages for each mode: \texttt{Bias} mode~(shift \texttt{Q$_\text{BEV}$}) demonstrates superior performance in perception-related metrics (NDS, mAP, and Collision Rate), while \texttt{TF} mode~(shift \texttt{F$_\text{BEV}$}) more effectively improves planning similarity. These results suggest that Query Shift better enhances perception capabilities, whereas Feature Shift more significantly benefits planning performance.
The observed performance improvements suggest that shift \texttt{Q$_\text{BEV}$} enhances perception by facilitating more effective extraction of critical visual features, whereas shift \texttt{F$_\text{BEV}$} does not significantly improve real-time perception but instead provides valuable prior knowledge to support long-term planning decisions.

The operation time for VAD and NetRoller models tested on the v1.0-mini subset with NVIDIA RTX 4090 is listed in Table~\ref{tab:time}. Besides, the time consumption of \texttt{VAD~base$\star$} is tested aside GM, acting as a baseline to evaluate the time consumption of information distribution. 
Experimental results demonstrate that the \texttt{C$_1$} mode achieves a 98\% reduction in information collection latency while introducing no perceptible computational overhead to the SM.
\TODO{This reduction in computational cost can substantially decrease the computing resources required in edge computing environments, while the frame rate can be further optimized through a receding-batch processing strategy.}

\begin{table}[]
    \centering
    \caption{Comparison of time consumption}
    \begin{tabular}{c|c|c}
    \toprule
        Model & Collection and Tanslation\,(s)$\downarrow$ & Distribution\,(s)$\downarrow$ \\ \hline
        \rowcolor[HTML]{EFEFEF} 
        {VAD base}$\star$& - & 0.1457  \\ \hline
        Bias-C$_0$& 3.6275$\pm0.4860 $  & 0.1410   \\
        Bias-C$_{\text{-}1}$& 3.5558$\pm0.4570  $   & 0.1372  \\
        Bias-C$_1$& 0.0667$\pm0.0074 $ & 0.1451 \\ \hline
        TF-C$_0$& 3.5445$\pm0.4585 $  & 0.1391  \\
        TF-C$_{\text{-}1}$& 3.5365$\pm0.4576 $  & 0.1382  \\
        TF-C$_1$& 0.0661$\pm0.0103 $    & 0.1491  \\
    \bottomrule
    \end{tabular}
    \label{tab:time}
\end{table}

\subsection{Frame-Wise Comparison}

\begin{figure}[ht!]
    \centering
    \includegraphics[width=\linewidth]{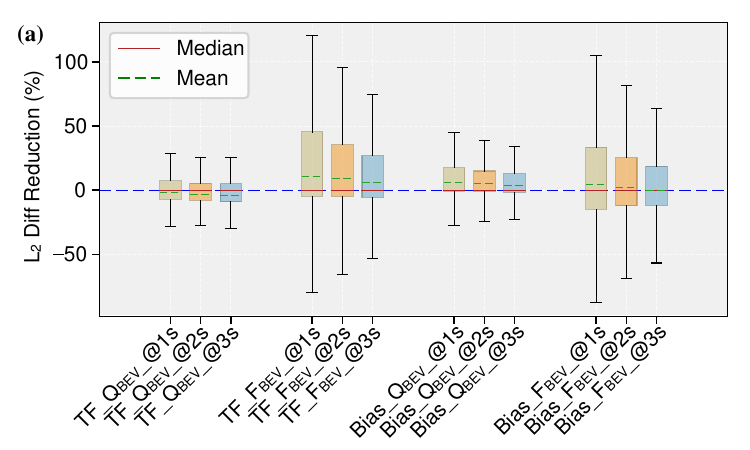} 
    \includegraphics[width=\linewidth]{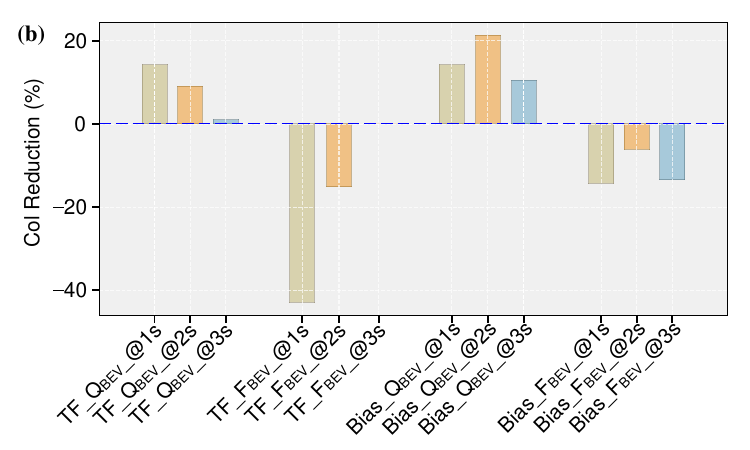}
    \vspace{-1em}
    \caption{Frame-wise analysis: (a) illustrates the increase in similarity across each scenario within the test set. (b) shows the average reduction of the collision rate within each horizon. All metrics in our evaluation framework are designed such that higher values indicate better performance.}
    \label{fig:instancewise}
\end{figure}

To further analyze the robustness and stability when incorporating VLMs into the system, we conduct a frame-wise statistical comparison. The metrics used include:
\begin{itemize}
    \item \textbf{Portion with Improvement}: This metric quantifies SM performance improvement across each frame. The positive portion of the distribution in Fig.~\ref{fig:instancewise}\,(a) reflects these enhancements. \TODO{This metric also reflects the efficiency and robustness of each method on extracting valuable information, contributing to the generation of more human-like and safer trajectories without introducing significant drawbacks.}
    \item \textbf{Maximum Withdrawal}: This metric evaluates the degradation of worst-case performance across frames, assessing whether the VLM-introduced information adversely affects the specialized model. The metric is quantified by the lower bound of linebox bars shown in Fig.~\ref{fig:instancewise}\,(a).
    \item \textbf{Average Improvement}: This metric quantifies the mean performance improvement per frame, as represented by the green dashed reference line in Fig.~\ref{fig:instancewise}\,(a).
\end{itemize}
The improvement of planning discrepancy between model output and ground truth is measured by Cartesian Distance~($L_2$ norm), calculated by $\Delta L_2=L_{2,base} - L_{2,exp}$, where $L_{2,base}$ represents planning discrepancy of baseline model and $L_{2,exp}$ represents planning discrepancy of our experimental GM enhanced system.
A positive $\Delta L_2$ value indicates reduced discrepancy between the model's planning decisions and human expert behavior, with larger values corresponding to better alignment.
Based on the results shown in Fig.~\ref{fig:instancewise}\,(a), it is evident that the \texttt{Bias} mode achieves the highest first quartile value, which represents the \textit{Portion with Improvement}, and the smallest \textit{Maximum Withdrawal}. Additionally, the \texttt{TF} mode demonstrates a significant increase in the \textit{Average Improvement}, but this comes with a lower \textit{Portion with Improvement} and a greater \textit{Maximum Withdrawal}. This phenomenon indicates that while the \texttt{TF} mode enhances performance for some frames, it detracts from others.
Advantages of \texttt{Bias} mode are cross-validated by Fig.~\ref{fig:instancewise}\,(b), demonstrating its ability to significantly reduce collision rates. 
In contrast, the \texttt{TF} mode increases collision risks.
This discrepancy is likely due to the lower flexibility of the Feature Shift mechanism. The higher-level instructions may cause the planning module to focus primarily on a specific target, potentially neglecting other important information, such as obstacles on the path. This could also be a contributing factor to the relatively high collision rate observed in Table~\ref{tab:main}.
\TODO{The result of framewise analysis demonstrates that training with latency and information and translating with learnable padding embedding can effectively prevent erroneous or misleading information from negatively impacting the SM.}

\subsection{Ablation Study} 

\begin{figure*}[ht!]
    \centering
    \begin{minipage}[t]{0.32\linewidth}
    \centering
    \includegraphics[width=\linewidth]{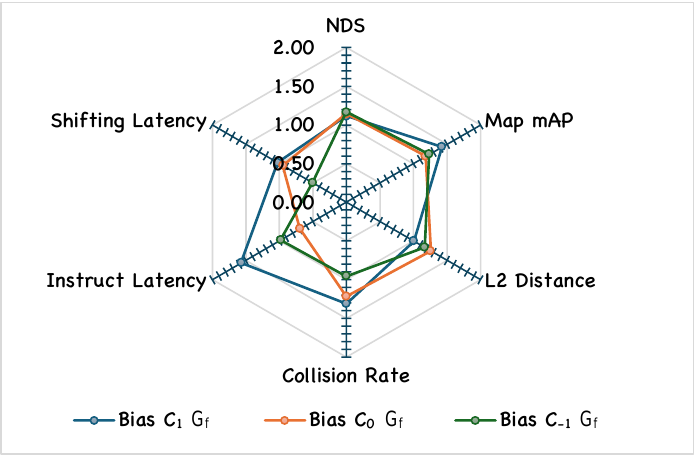}
    \centering
    \footnotesize{(a.1) Collection modes w/o \texttt{PE$_{l}$} for \texttt{Bias}}
    \vspace{1em}
    \end{minipage}
    \begin{minipage}[t]{0.32\linewidth}
    \centering
    \includegraphics[width=\linewidth]{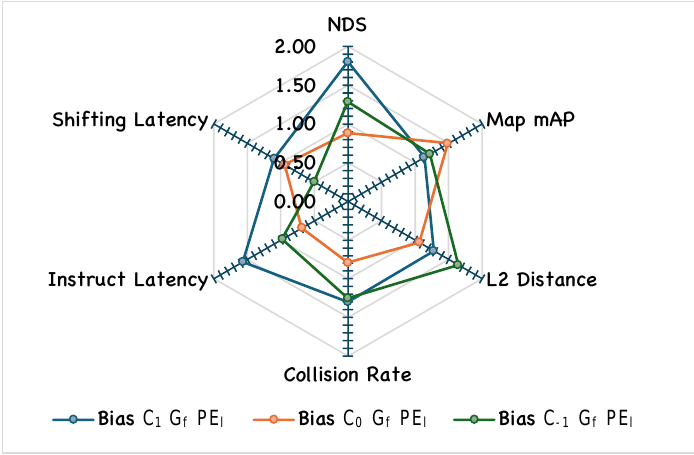}
    \centering
    \footnotesize{(a.2) Collection modes with \texttt{PE$_{l}$} for \texttt{Bias}}
    \end{minipage}
    \begin{minipage}[t]{0.32\linewidth}
    \centering
    \includegraphics[width=\linewidth]{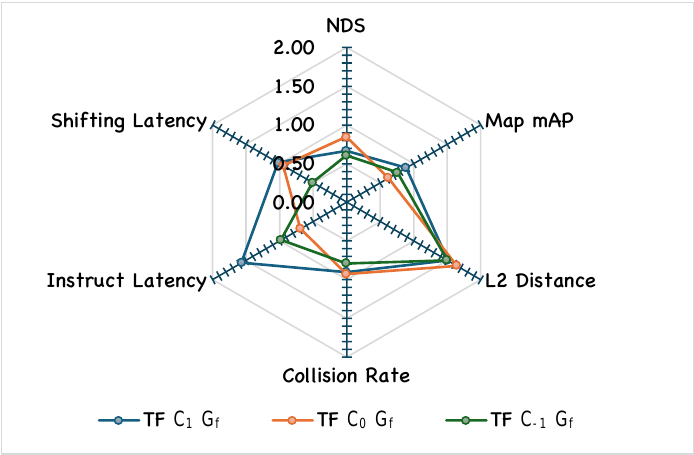}
    \centering
    \footnotesize{(a.3) Collection modes w/o \texttt{PE$_{l}$} for \texttt{TF}}
    \end{minipage}
    \begin{minipage}[t]{0.32\linewidth}
    \centering
    \includegraphics[width=\linewidth]{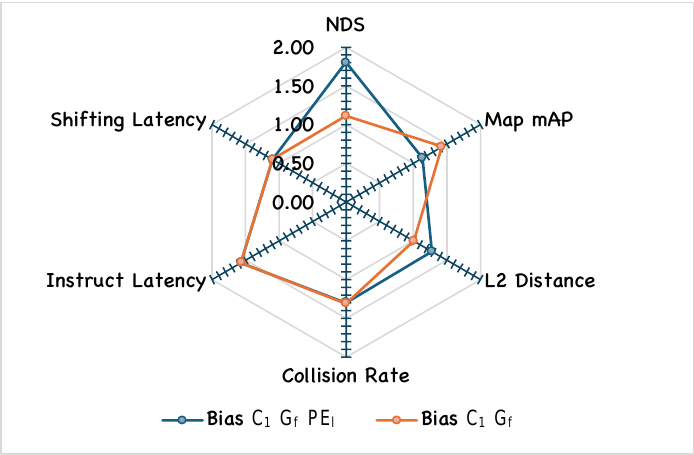}
    \footnotesize{(b.1) Effect of \texttt{PE$_{l}$} for \texttt{Bias}} 
    \vspace{1em}
    \end{minipage}
    \begin{minipage}[t]{0.32\linewidth}
    \centering
    \includegraphics[width=\linewidth]{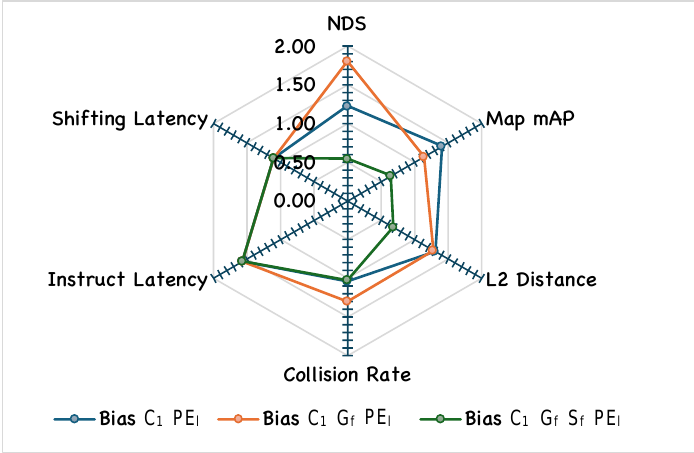}
    \footnotesize{(b.2) Effect of $G_f$/$S_f$ for \texttt{Bias}}
    \end{minipage}
    \begin{minipage}[t]{0.32\linewidth}
    \centering
    \includegraphics[width=\linewidth]{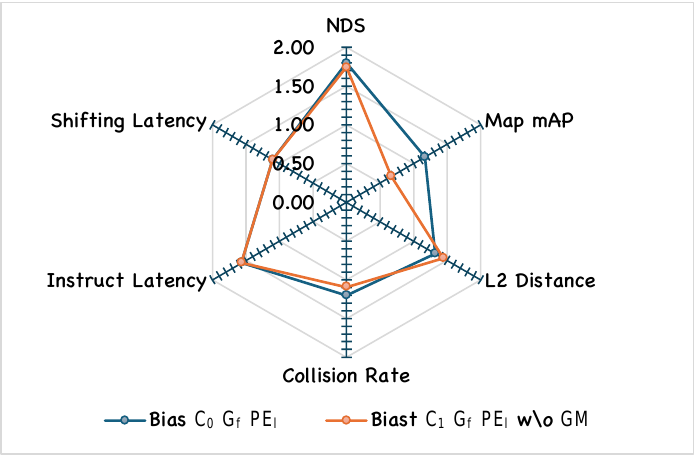}
    \footnotesize{(b.3) Inference w/o GM for \texttt{Bias}}
    \end{minipage}    
    \begin{minipage}[t]{0.32\linewidth}
    \centering
    \includegraphics[width=\linewidth]{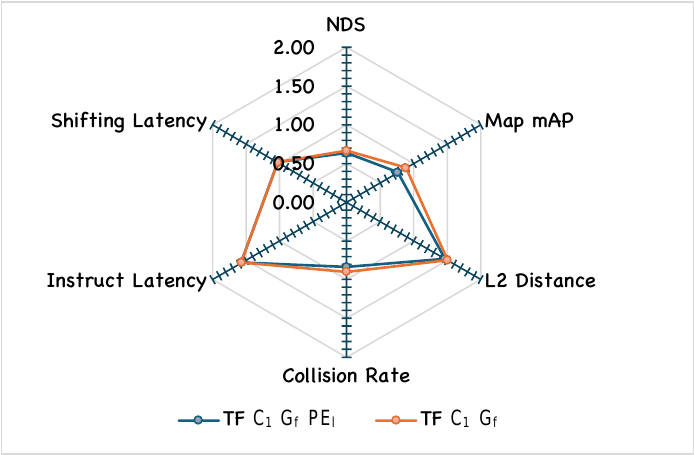}
    \footnotesize{(c.1) Effect of \texttt{PE$_{l}$} for \texttt{TF}}
    \end{minipage}
    \begin{minipage}[t]{0.32\linewidth}
    \centering
    \includegraphics[width=\linewidth]{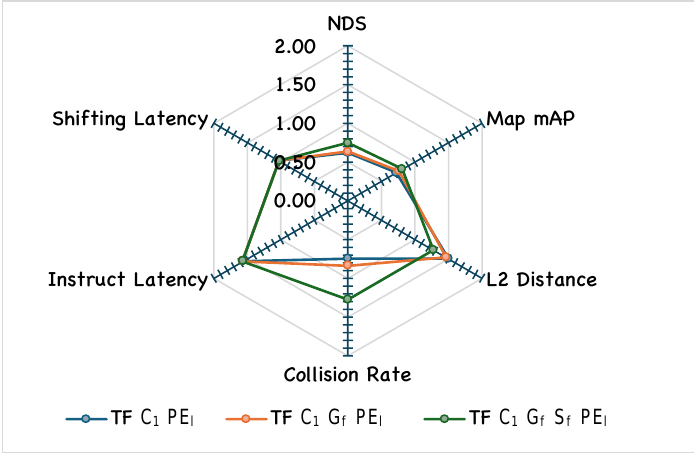}
    \footnotesize{(c.2) Effect of $G_f$/$S_f$ for \texttt{Bias}}
    \end{minipage}
    \begin{minipage}[t]{0.32\linewidth}
    \centering
    \includegraphics[width=\linewidth]{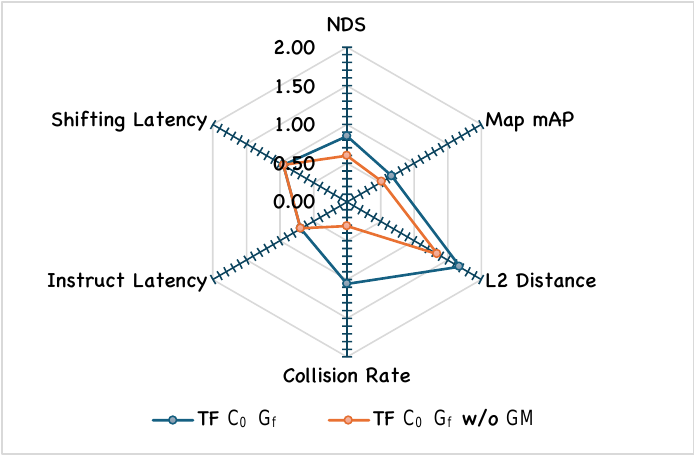}
    \footnotesize{(c.3) Inference w/o GM for \texttt{TF}}
    \end{minipage}
    \caption{Results of the ablation studies are presented. The first row compares the collection mechanisms based on Bias, with and without le and TF modes. The second and third rows analyze the effects of le, the impact of the fine-tuning portion, and the contribution of GM in both Bias and TF modes, respectively.
    }    
    \label{fig:ablation}
\end{figure*}

\begin{figure}[ht]
    \centering
    \begin{minipage}[t]{0.75\linewidth}
    \centering
    \includegraphics[width=\linewidth]{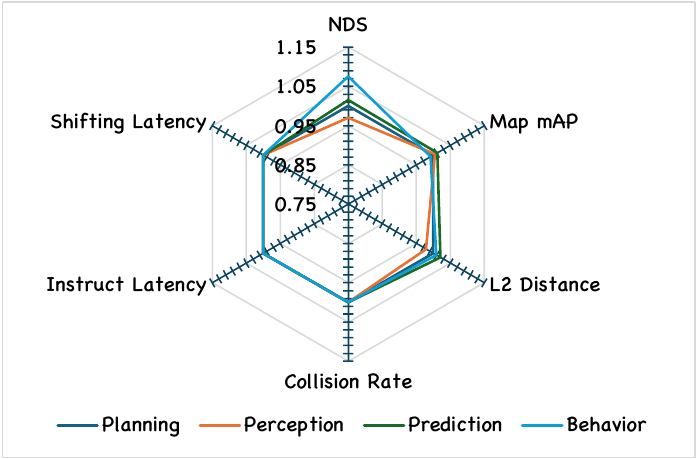}
    \footnotesize{(a) QA type comparison for \texttt{Bias}}
    \end{minipage}
    \centering
    \begin{minipage}[t]{0.75\linewidth}
    \centering
    \includegraphics[width=\linewidth]{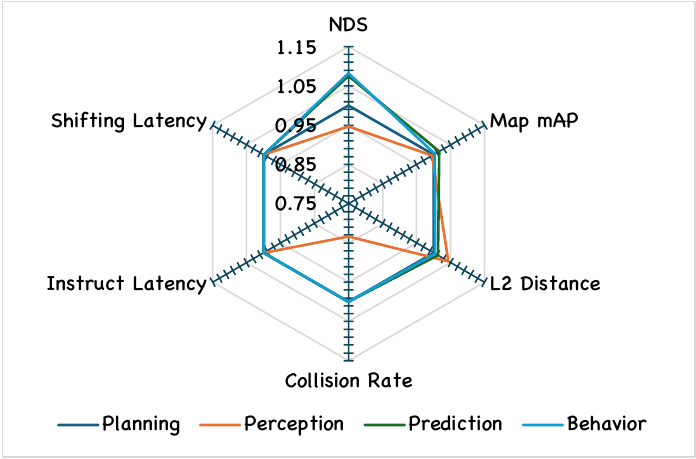}
    \footnotesize{(b) QA type comparison for \texttt{TF}}
    \end{minipage}
    \caption{Comparison of different QA types for \texttt{Bias} and \texttt{TF} modes. In each figure, planning related QA serves as the baseline and the axis represents the advantage value of other QA type. }
    \label{fig:ablation_qa}
\end{figure}

To assess the impact of each component on the performance of the system, we conduct ablation studies and multi-dimensional analysis. To facilitate a more intuitive comparison of data changes across various dimensions, the data were converted into normalized advantage values.
The advantage value presented in the radar diagrams is calculated by
\begin{equation}
    \text{Advantage} = 1 \pm \frac{M-M_{base}}{\max(M_{exp})-\min(M_{exp})},
\end{equation}
where $M$ represents the value of a metric, $M_{base}$ represents the metric value of the baseline model, $M_{exp}$ represents the set of metric values measured across all configurations, $\pm$ indicates that opposite sign values are used for deviation ratios to ensure consistent metric interpretation, where higher values always represent better performance across all measurements.

\paragraph{Information Collection Strategy}
In Fig.~\ref{fig:ablation}\,(a.1), (a.2) and (a.3), the information extraction methods of large language model tokens, including the first token from all layers~(\texttt{C$_1$}), all tokens from all layers~(\texttt{C$_0$}), and all tokens from the last layer~(\texttt{C$_{\text{-}1}$}), are compared. As can be seen from the figure, all the above methods show improved planning effects compared to the baseline. 
Our proposed first-layer all-token approach achieves dual advantages in reducing computational burden and keeping higher detection precision, safety, and human similarity. This demonstrates the effectiveness in balancing the computational efficiency of the NetRoller adaptor.

\paragraph{Effect of Positional Layer Embedding}
In the "first token all layer"~(\texttt{C$_1$}) information collection mode, the difference between collecting with and without layer embeddings, in both \texttt{Bias} and \texttt{TF} mode, is compared and illustrated in Fig.~\ref{fig:ablation}\,(b.1) and (c.1). From the results shown by the \texttt{Bias} mode, it can be observed that adding layer embeddings significantly improves the NDS and $L_2$ Distance metrics related to the actual description of the text. This improvement indicates that adding $PE_{layer}$ helps the model better identify valuable contextual information from the upper-layer variables.
This module has little impact on the \texttt{TF} mode. 
A possible reason is that \texttt{TF} mode equips higher-dimensional $E_R$, which enhances its information extraction capability.
The higher-dimensional $E_R$ in the \texttt{TF} mode may already provide sufficient context and structure, making the additional benefit of $PE_{layer}$ less noticeable.

\paragraph{Effect of Joint Finetuning}
In the following experiment shown in Fig.~\ref{fig:ablation}\,(b.2) and (c.2), we focused on comparing the impact of the parameters involved in the joint training. 
This experiment was primarily used to analyze which parameters need to be fine-tuned for NetRoller to be effective.
The comparison in Fig.~\ref{fig:ablation}\,(b.2) shows that for the \texttt{Bias} mode, fine-tuning of the downstream query is necessary. This is because the query needs to change from an absolute value to a relative value based on the bias. For the \texttt{F$_\text{BEV}$} mode using \texttt{TF}, since it offers the higher-level instructions pass the \texttt{Key}, \texttt{Value}, and 
learns the parameters of \texttt{Transformer}.
It directly contributes to the planning results.

This experiment evaluates whether GMs can enhance real-time SM reasoning performance for perception and planning tasks. To isolate the impact of model reasoning while simulating practical deployment conditions, we masked all input features to the GM while preserving its output connections to downstream components. This configuration tests the system's robustness against potential noise from irrelevant model outputs. As demonstrated in Fig.~\ref{fig:ablation}\,(b.3) and (c.3), incorporating large-scale models yields significant performance gains across multiple metrics, with particularly notable improvements in key indicators.
Surprisingly, the \texttt{Bias} configuration maintains competitive performance across most metrics, even without GM integration.
\TODO{This finding indicates that, by providing high-level conditional information, the GM allows the SM to focus on its reasoning more effectively, reducing attention to irrelevant details. As a result, the SM demonstrates improved performance even when deployed independently under average conditions in general scenarios.}

\TODO{
Further investigation into the impact of different QA types is primarily conducted on the nuScenes-mini dataset. Following the taxonomy used in DriveLM, QA types are categorized into four groups: planning, perception, prediction, and behavior. As illustrated in Fig.~\ref{fig:ablation_qa}, behavior and prediction related QA pairs exhibit a more significant positive effect in both \texttt{Bias} and \texttt{TF} modes. In contrast, perception-related QA pairs unexpectedly demonstrate a negative impact on model performance.
Upon analyzing the dataset, it becomes apparent that perception-related QA pairs often contain planning irrelevant or redundant information—such as vehicle colors, types, or other loosely related scenario descriptions. This inclusion of extraneous details may cause the model to focus on irrelevant features, thereby impairing its effectiveness. Conversely, planning and prediction related QA pairs tend to provide more straightforward information for trajectory planning, such as road conditions, traffic signs, and anticipated behaviors of surrounding agents. Behavior-related QA pairs, in particular, are closely aligned with the ego vehicle's planning task, offering future behavioral cues with an appropriate level of detail.
However, it is important to note that the QA categorization in DriveLM is not strictly limited to one category, and individual QA pairs may span multiple categories. Future research using more clearly defined taxonomies and larger datasets is necessary to thoroughly validate the differentiated effects of QA types on planning performance.
}

\section{Conclusion}
In this paper, we propose NetRoller, a novel adaptor that is designed to efficiently integrate general intelligence from GM into the specialized autonomous driving system. Specifically, we formalize the novel mechanisms into three stages, encompassing information collection, translation, and distribution to utilize informative features across both systems effectively. We empirically investigate Query Shift and Feature Shift as two strategies for distributing the information for downstream applications. Through a series of comparative experiments, the specialized model achieves a 16.71\% reduction in collision rate with Query Shift, and a 12.46\% improvement in trajectory similarity with Feature Shift. 

\TODO{Compared to the traditional modularized planning framework, the GM acquires human-like reasoning capabilities with fewer hand-crafted rules and a larger scale general knowledge. It can be regarded as an additional high-level abstraction layer that considers not only planning related information but also incorporates human-like reasoning and common sense priors. This supplementary layer of information supports the SM in making more informed long-term decisions within the coarse-to-fine planning paradigm~\cite{planscope}.
}

Notwithstanding that the backbone LLM we employed for the specialized driving model, LLaMA 2~\cite{llama2}, falls short in performance and robustness when compared to closed-source LLMs, the proposed methodology still demonstrates significant potential. Particularly, further optimization of the training strategy and model structure could enhance its performance to a greater extent. \TODO{For instance, the proposed framework could serve as a bridge between the modules in the autonomous driving pipeline. It can deliver more flexible and informative messages while preserving the original sensor frame rate to avoid redundant sensor signal process.}
Besides, to fully capitalize on the mechanisms proposed in this work, several avenues for enhancement can be pursued. These include refining the dataset and the QA format to better suit the needs of the framework, exploring the application of asynchronous neural networks in a broader range of tasks to validate their efficacy, and considering the integration of more robust GMs to enhance overall performance. 

\bibliographystyle{IEEEtran}
{\small \bibliography{ref}}

\end{document}